\pdfoutput=1

\documentclass[11pt]{article}

\usepackage{acl}

\usepackage{times}
\usepackage{latexsym}
\usepackage{amsmath}
\usepackage{tabularx}

\usepackage[T1]{fontenc}

\usepackage[utf8]{inputenc}

\usepackage{microtype}

\usepackage{inconsolata}
\usepackage{booktabs}
\usepackage{pdflscape}

\title{Language Resources for Dutch Large Language Modelling}

\author{Bram Vanroy \\
        KU Leuven \\ Oude Markt 13 \\ Leuven, Belgium \\ \texttt{bram.vanroy@kuleuven.be}
}

\begin{document}
\maketitle
\begin{abstract}

Despite the rapid expansion of types of large language models, there remains a notable gap in models specifically designed for the Dutch language. This gap is not only a shortage in terms of pretrained Dutch models but also in terms of data, and benchmarks and leaderboards. This work provides a small step to improve the situation. First, we introduce two fine-tuned variants of the Llama 2 13B model. We first fine-tuned Llama 2 using Dutch-specific web-crawled data and subsequently refined this model further on multiple synthetic instruction and chat datasets. These datasets as well as the model weights are made available. In addition, we provide a leaderboard to keep track of the performance of (Dutch) models on a number of generation tasks, and we include results of a number of state-of-the-art models, including our own. Finally we provide a critical conclusion on what we believe is needed to push forward Dutch language models and the whole eco-system around the models.
\end{abstract}

\section{Introduction}
\label{sec:introduction}

Generative Large language models (LLMs) have experienced rapid advancements in recent years, particularly thanks to their transformer-based architecture \cite{vaswani2017attention}, improvements in computational processing efficiency and hardware capabilities, and the collection and creation of massive datasets. However, these models most often, with a few exceptions, focus on English with minor attention to other languages. While this is understandable due to the international status that English has as well as the availability of English data, a significant gap remains in the development of specialised language models for languages that are not as widely represented. Dutch, being a language spoken by around 24 million people according to the Dutch ``Taalunie'',\footnote{\url{https://taalunie.org/informatie/24/feiten-cijfers}} is one such language that remains underrepresented in the realm of language modelling. A broader collection of Dutch LLMs will push forward academic and industrial research on automated language generation while also paving the way for non-expert users to use generative artificial intelligence (AI) tools in their own mother tongue. Importantly, in terms of accessibility and availability, making sure that the models are open-source encourages ethical practices and accountability by allowing for transparent scrutiny of the models' functionality and behaviour. However, as will become obvious in the remainder of this paper, large language models are an end product that can only be achieved with appropriate datasets. Furthermore, incremental improvements in these models can only be achieved when sufficient evaluation methods can be employed. 

Against this backdrop, this study introduces a number of language resources to accelerate research. This includes a number of instruction datasets, two finetuned two fine-tuned variants of the Llama~2 \cite{touvronLlamaOpenFoundation2023} 13B model, and a leaderboard. In terms of datasets we publish translated variants of a number of popular instruction datasets: Dolly, Quora, Stack Overflow and Alpaca, which we translated to Dutch with \texttt{gpt-3.5-turbo}. We also release code for the automatic translation of datasets with either local models (that are present on the Hugging Face hub) or with Azure's OpenAI services. The models that we trained, with limited compute, were inspired by the difficulties that we experienced to chat with Llama~2. We found that the model often would use English words, fall back to English after a few sentences, or just read as a poorly English-to-Dutch translation. The first model is trained on causal language modelling, also known as regular text completion, of web-crawled, Dutch text. The second version builds on the former by fine-tuning further on synthetic (translated) datasets that contain instructions and conversational data. Both models are trained on a context window of 4096 tokens. To facilitate broader research, both our models are made available as well as quantised versions of the chat variant for efficient deployment.\footnote{\url{https://huggingface.co/collections/BramVanroy/finetuned-llms-for-dutch-64f99dddbb86c0fa80846f89}} Additionally, the synthetic datasets are released (Sec.~\ref{sec:datasets}). Lastly, we also populated and maintain a leaderboard for Dutch generative language modelling (Sec.~\ref{sec:leaderboard}), where we recorded results of many state-of-the-art models as well as our own. While benchmark results of our model, as well as other Dutch-specific models, are modest compared to the base model, we envision that the resulting models, trained with limited compute, can serve as a baseline for the development of Dutch models, and that the dataset and model descriptions in this paper as well as the discussions can ignite the training of better models for this language.

\section{Datasets}
\label{sec:datasets}

We release two models (Sec.~\ref{sec:models}): one general text completion model, and another chat variant that can follow instructions. The first version is trained on a version of the multilingual C4 corpus, which is based on the CommonCrawl web-crawled corpus \cite{raffelExploringLimitsTransfer2020}. This specific version\footnote{\url{https://huggingface.co/datasets/yhavinga/mc4_nl_cleaned}} contains a cleaned version of the Dutch partition of the mC4 corpus, where, among other processing techniques, Dutch obscene and bad words are removed, and too short (\(<3\) words) or too long (\(> 250\) characters) sentences are removed, yielding a high-quality, safe-for-work corpus. We trained for one epoch on the \texttt{tiny} partition which contains around 2B words. 

The second model, with chat capabilities, builds on the former by fine-tuning further on translated instruction datasets, which we make publicly available.\footnote{\url{https://huggingface.co/datasets/BramVanroy/dutch_chat_datasets}} All datasets were translated with \texttt{gpt-3.5-turbo} to Dutch. Note however that none of these automatic translations were manually verified and may contain nonfactual, inaccurate and disfluent translations. For transparency, we give the translation cost but note that these translations were made at a time where the cost of the API may have been more expensive than it is now so the numbers may not correspond to today's cost. For details on the translation process and prompts, see Appendix~\ref{app:translation}. We release the scripts to translate datasets on the Hugging Face hub on GitHub, allowing the use of either Azure's OpenAI services or open-source models on the Hugging Face hub to translate or answer datasets.\footnote{\url{https://github.com/BramVanroy/dutch-instruction-datasets}}

\paragraph{Dolly 15k} Dolly 15k \cite{DatabricksBlog2023DollyV2} is a human-created dataset of instructions, where the Databricks company incentivized their employees to create prompt-response pairs in behavioural categories such as brainstorming, classification, closed and open question answering, generation, information extraction and summarization, inspired by OpenAI's InstructGPT paper \cite{ouyang2022TrainingLanguageModels}. In addition, open-ended free form questions were also included. All data was manually generated without the use of generative assistance but the data creators could use Wikipedia as a reference. Translating this dataset of 15,000 samples cost around \$19.38. The translated dataset is available on the Hugging Face hub.\footnote{\url{https://huggingface.co/datasets/BramVanroy/dolly-15k-dutch}}

\paragraph{Quora} Quora is a user-centered questions-and-answers website. Authors of Baize \cite{xu2023BaizeOpenSourceChat} used questions posed on that platform as a ``seed'' question, and then used \texttt{gpt-3.5-turbo} to create a self-chat, multi-turn conversation, where GPT~3.5 generates a conversation with itself. This resulted in a conversational dataset of around 55.000 samples, which we translated to Dutch for \$135.65. The translations are available at this URL.\footnote{\url{https://huggingface.co/datasets/BramVanroy/quora-chat-dutch}}

\paragraph{Stack Overflow} The Stack Overflow website is part of the Stack Exchange network of Q\&A sites, specifically dedicated to programming questions. Similar to the Quora chat dataset, the Baize authors \cite{xu2023BaizeOpenSourceChat} used questions posted on the Stack Overflow platform as seed questions, and then prompted \texttt{gpt-3.5-turbo} to converse with itself in multiple turns. The 56,964 translated conversations cost \$133.60 to translate and are available on the Hugging Face hub.\footnote{\url{https://huggingface.co/datasets/BramVanroy/stackoverflow-chat-dutch}}

\paragraph{Alpaca} Researchers at Stanford also followed the InstructGPT \cite{ouyang2022TrainingLanguageModels} to generate instructional data and made their resulting dataset of around 52,000 samples publicly available \cite{alpaca}. These single-turn instructions were generated with OpenAI's \texttt{text-davinci-003} model. The researchers claim that their dataset is diverse and covers many different topics, making it suitable to train open-source instruction following models. This dataset was ``cleaned'' from halluciniations and empty outputs by user \texttt{yahma} on the Hugging Face hub, which we used as a starting point for our translation to Dutch.\footnote{\url{https://huggingface.co/datasets/yahma/alpaca-cleaned}} The translated dataset, which costs \$57.99 to generate, therefore consists of 51,712 single-turn conversations with instructions, optional input, and output. It is available on the Hugging Face hub.\footnote{\url{https://huggingface.co/datasets/BramVanroy/alpaca-cleaned-dutch}}

\section{Training Procedure}

To create our models we were severely compute restrained. The goal was to provide models that were finetuned for Dutch, but that were otherwise identical to Llama~2 in terms of usage and capabilities. The goal was mostly a chat model that was capable of maintaining a conversation in Dutch, whereas we found that the original Llama~2 models have difficulties with that. Specifically we wanted to finetune the model in full coverage of its initial context length of 4096 so that the models can be used for large pieces of texts or long conversations. Note that our models were initially trained earlier this year (around mid August, 2023) and are therefore already ``out-dated'' in the fast-moving model landscape. Since then, many new models have been introduced, which we include in our leaderboard (cf.~Sec.~\ref{sec:leaderboard}).

Due to the computational constraints we decided to finetune on QLoRA \cite[i.e. 4-bit quantization with LoRA adapter layers;]{dettmers2023qlora}. We targeted only the \texttt{q\_proj} and \texttt{v\_proj} linear layers with LoRA. If more compute were available, we would recommend to do a full finetune of the model or at least include all linear layers in the LoRA. After all, we are trying to get the model to realign with a different main target language, which seems an important part of the whole model and not just of small adapters. We used flash attention to train our models \cite{dao2022flashattention}.

The finetuned base model (\texttt{llama2-13b-ft-mc4\_nl\_cleaned\_tiny}) was finetuned for only 120 GPU hours in total on four nodes of four NVIDIA A100 80GB GPUs each. In that time it trained one epoch on the mc4\_nl\_cleaned dataset (tiny partition). Hyperparameters of the base model can be found here.\footnote{\url{https://huggingface.co/BramVanroy/llama2-13b-ft-mc4_nl_cleaned_tiny\#training-hyperparameters}}. The chat model (\texttt{Llama-2-13b-chat-dutch}) was finetuned for only around 55 GPU hours in total on four NVIDIA A100 80GB GPUs. That means that the four datasets (Dolly, Alpaca, Stack Overfloa and Quora) above were trained on for two epochs. The main intention of this step was to teach the model the conversational nature and to follow the chat template. The finetuned model's training hyperparameters can be found on its model's card.\footnote{\url{https://huggingface.co/BramVanroy/Llama-2-13b-chat-dutch\#training-hyperparameters}} TheBloke, a beacon in the open-source community for model quantization, has also released quantized versions of the chat model.\footnote{\url{https://huggingface.co/models?search=thebloke\%20chat\%20dutch}}

\section{Benchmarks}
While Dutch benchmarks for language-related tasks exist (cf. the collection in DUMB, \cite{de-vries-etal-2023-dumb}), such benchmarks are typically aimed at (fine-tuning) encoder models for a classification task. Benchmarking generative models on such classification tasks is possible with zero or multi-shot approaches and suitable prompting, but we present results on such benchmarks in other work \cite{benchmarking2023delanghe}.

Benchmarks specifically for generative, Dutch large language models do currently not exist to the best of our knowledge. Luckily, some efforts have been made to widen the reach of existing evaluation benchmarks of English to other languages. \citet{dac2023okapi} provide an evaluation harness, based on EleutherAI's harness \cite{eval-harness}, to evaluate multilingual models. In their work, they provide translations of a number of common benchmarks for English LLMs that require the model to choose the correct option out of multiple choice questions. Below are the descriptions of the original benchmarks (English) but we use the translated versions to Dutch by \citet{dac2023okapi} in our evaluations.

\paragraph{ARC}

The AI2 Reasoning Challenge \cite[ARC]{clark2018think} is a challenge set for complex question answering. ARC focuses on natural science questions from grade-school exams that require deep reasoning and knowledge. It comprises a Challenge Set and an Easy Set, with the former containing questions that both retrieval-based and word co-occurrence algorithms fail to answer correctly. Models are expected to select the right answer out of a number of given options.

\paragraph{HellaSwag}

HellaSwag \cite{zellers2019hellaswag} is a dataset for evaluating commonsense reasoning through natural language inference. It presents challenging scenarios requiring models to choose the most plausible continuation from multiple choices, focusing on understanding and predicting everyday activities. Despite being relatively straightforward for humans, the benchmark still poses challenges for large language models. This benchmark aims to push the envelope in natural language understanding and commonsense reasoning, highlighting the gap between current machine learning capabilities and human-like understanding.

\paragraph{MMLU}

As the name implies, the Massive Multitask Language Understanding benchmark \cite[MMLU]{hendrycks2021measuring} is designed to measure the multitask accuracy of text models across 57 diverse subjects, ranging from elementary mathematics to professional medicine in varying degrees of difficulty. It is specifically designed to assess a model's depth and breadth of academic and professional understanding, focusing on real-world language comprehension. The goal is to identify a model’s strengths and weaknesses in various disciplines, emphasizing the necessity for models to have extensive world knowledge and problem-solving capabilities.

\paragraph{TruthfulQA}

The TruthfulQA benchmark \cite{lin2022truthfulqa} assesses the truthfulness of language model-generated answers to questions. It is therefore focused on facts and world knowledge and less on reasoning. The benchmark includes challenges from 38 categories such as health, law, finance, and politics, but also specifically fact-focused ones like conspiracies or fiction. These questions are designed to elicit responses that reflect common misconceptions or false beliefs. The objective is for models to avoid reproducing these falsehoods, which they may have learned from human texts, which naturally contain human biases as well.

\paragraph{Leaderboard}
\label{sec:leaderboard}
To have a clear focus on Dutch and to be able to benchmark new models quickly, we created a new leaderboard\footnote{\url{https://huggingface.co/spaces/BramVanroy/open_dutch_llm_leaderboard}} that is separate from the initial multilingual leaderboard\footnote{\url{https://huggingface.co/spaces/uonlp/open_multilingual_llm_leaderboard}}. We also extend the leaderboard with new information and links to all the models. We keep track of information such as how the model was trained (pretrained, fine-tuned, instruction-tuned, preference-tuned with reinforcement learning) and whether it was specifically trained for Dutch in the pretraining or fine-tuning stage, or not. Finally we also list the model sizes. This is especially useful because we have benchmarked already a wide range of decoder models, including state-of-the-art ones that are mostly focused on English, like Zephyr \cite{tunstall2023zephyr} but also older models that were specifically pretrained on Dutch, like GPT2 Dutch.\footnote{\url{https://huggingface.co/yhavinga/gpt2-medium-dutch}}

While we do not have the compute power to accept model name submissions to an automated system that would continuously benchmark any submitted model, we are eager to work with the community to add new models. To ensure fair evaluations, we will run benchmarks on new models upon request but only if there is a technical report or paper related to the model, and only if the model and data has been transparently described. We understand that data can be sensitive and not publicly available for a number of reasons like privacy or copyright, but it should at least be described clearly in terms of domain, size and content. Model weights must be publicly available for the model to be added to  the leaderboard. Model creators can request us to benchmark their model and add it to the overview page by creating a new thread on this leaderboard page: \url{https://hf.co/spaces/BramVanroy/open_dutch_llm_leaderboard/discussions}.

\section{Benchmark Models}
\label{sec:models}

In addition to our introduced models, we also benchmark a number of models on the described benchmarks for Dutch. Due to the quick rise in performance of models, we also include models that were not specifically trained on Dutch because their multilingualism seems to increase, too.

\paragraph{Pretrained Dutch GPT models} Yeb Havinga has pretrained a number of models on Dutch. These are either GPT-2 or GPT-Neo based in terms of architecture but they have been trained from-scratch on Dutch data only, specifically on a cleaned version of the Dutch mC4 corpus that was described above and that we also used to finetune Llama~2. The models range in size between 125M to 1.3B parameters: \texttt{gpt-neo-125m-dutch}, \texttt{gpt2-medium-dutch} (355M), \texttt{gpt2-large-dutch} (774M) and \texttt{gpt-neo-1.3b-dutch}. All models can be found on Havinga's model page.\footnote{\url{https://huggingface.co/yhavinga}} The models were not tuned for following instructions or chat.

\paragraph{Llama 2} Llama~2 \cite{touvronLlamaOpenFoundation2023}, developed by Meta AI, is the continuation of Llama and has played a large role in large companies distributing the weights of their large language models.\footnote{\url{https://huggingface.co/meta-llama}} They have released 7B, 13B and 70B variants of their models, which come in both the base model variant and a chat model variant, but a 34B model exists but was not released. Due to computational limitations, we only include 7B and 13B variants. Models were trained on a mix of publicly available online data totalling 2T tokens, although details are scarce. The base models were finetuned on newly collected, high-quality instruction and chat data. They find that a smaller but high-quality set of annotations performs much better than mullions of data points. They collected 27,540 conversations but the dataset is not publicly available. In a second step, the models were also further aligned with human preference through reinforcement learning from human feedback \cite{christiano2023deep}, for which they collected another dataset where humans indicated which of two model outputs they preferred. This dataset was then used to train a reward model, which in turn served to optimise the chat models. Our own models have been described above. In brief, we started from the Llama~2 13B model, finetuned it on Dutch running text, and then finetuned the resulting model on synthetic Dutch chat datasets that we have publicly released. All Llama~2 based models have a default context window of 4096 tokens.

\paragraph{Orca-2} We include Orca~2 \cite{mitra2023orca}, which is a finetuned version of Llama~2. Through finetuning with specific details on diversity of tasks and data, the authors create what they call a ``cautious reasoner'' where the model has learnt to reason in multiple steps or strategize how to solve specific tasks. It is therefore an instruction tuned model that comes in different sizes. We include the 7B and 13B variants.\footnote{\url{https://huggingface.co/microsoft?search_models=orca}} They find that on English benchmarks, Orca~2 7B performs similarly to Llama~2 13B, and Orca~2 13B outperforms Llama~2 13B, often even outperforming Llama~2 70B.

\paragraph{Mistral} MistralAI released \texttt{Mistral-7B-v0.1}\footnote{\url{https://huggingface.co/mistralai/Mistral-7B-v0.1}}, which outperforms Llama~2 13B and Llama~1 34B on all English benchmarks \cite{jiang2023mistral}. It is trained from-scratch and therefore not based on Llama~2 or other existing models. Mistral is built with technical innovations such as sliding window attention so that attention can be paid to tokens outside of the context window of 8192 tokens that it was initially trained on. Unfortunately no information is known about the dataset that the model was trained on.

\paragraph{Zephyr} Zephyr is a chat model that has been created on the basis of Mistral, by creating a self-instruct dataset, supervised fine-tuning for instruction following, and preference optimisation \cite{tunstall2023zephyr}. \footnote{\url{https://huggingface.co/HuggingFaceH4/zephyr-7b-beta}} Especially interesting here, and different from previously listed models, is the usage of Direct Preference Optimization \cite[DPO]{rafailov2023direct} as a technique for aligning the model with expected results, specifically distilled DPO, where AI feedback from an ensemble of teacher models is used as preference data, to approximate what the best response is to a given question. This approach does not require human annotation and is therefore faster and likely cheaper than creating humanly annotated datasets. In the first step, Mistral was finetuned on a variant of UltraChat \cite{UltraChat}. For the second alignment step, they relied on the UltraFeedback dataset \cite{cui2023ultrafeedback} where different model completions were ranked by GPT-4 to indicate the preferred responses. The result of these steps is Zephyr. We specifically make use of Zephyr-beta, which was shown to outperform Llama~2 70B Chat on a variety of English benchmarks.

\paragraph{Neural Chat} Intel, the well-known chip manufacturer, developed ``Neural Chat'', a chat model based on Mistral.\footnote{\url{https://medium.com/intel-analytics-software/the-practice-of-supervised-finetuning-and-direct-preference-optimization-on-habana-gaudi2-a1197d8a3cd3}} In a first step, they finetuned Mistral on instruction following with the SlimOrca dataset \cite{SlimOrca}. For aligning the model better with human expectations, they also build and release a DPO dataset based on OpenOrca \cite{OpenOrca} that contains ``chosen'' and ``rejected'' answers to a given instruction.\footnote{\url{https://huggingface.co/datasets/Intel/orca_dpo_pairs}} The model will therefore learn to prefer the chosen answers and stay away from the rejected answers. In spirit, this resulting model\footnote{\url{https://huggingface.co/Intel/neural-chat-7b-v3-1}} is therefore very similar to Zephyr. This similarity is reflected in the results on English benchmarks, where both models perform on par with each other.

\paragraph{GEITje} Very recently, another Dutch-specific series of models was released.\footnote{\url{https://github.com/Rijgersberg/GEITje}} GEITje and its chat variant are based on the aforementioned Mistral model. The base model was trained on GigaCorpus-NL\footnote{\url{http://gigacorpus.nl/}}, which is a large collection of Dutch data containing, among others, legal texts, books, Dutch parts of Common Crawl, subtitles, news articles, and Twitter data. The training data also included the Dutch part of MADLAD-400 \cite{kudugunta2023madlad400}. In contrast to our Llama~2 models finetuned on Dutch, GEITje was created with full-finetuning (so no LoRA or other parameter-efficient strategies). GEITje Chat is a continuation of that model, further finetuned on conversational data, specifically on translations of the No Robots \cite{no_robots} dataset\footnote{\url{https://huggingface.co/datasets/Rijgersberg/no_robots_nl}} and of a portion of UltraChat \cite{UltraChat}\footnote{\url{https://huggingface.co/datasets/Rijgersberg/ultrachat_10k_nl}}. Unlike Zephyr and Neural Chat, no preference optimisation has been done.

\section{Results}
\label{sec:results}

In Table~\ref{tab:results} we present model results on the benchmarks. Note that the leaderboard is continuously updated with new models so for the latest updates, see \url{https://huggingface.co/spaces/BramVanroy/open_dutch_llm_leaderboard}. Best performance is highlighted in bold, but for representation purposes the results are limited to two decimals, which explains why some seemingly ties are not both in bold.

We can observe a number of highlights from these results. First of all, models with fewer than 6M parameters seem to perform much worse than others, even though they have all been pretrained on Dutch. However, looking more closely at these results we can see that this difference is especially noticeable in the few/multi-shot tasks. For TruthfulQA (a zero-shot task) the difference is smaller, where gpt2-medium-dutch is even outperforming Llama2 13B models. We hypothesize that this is not necessarily caused by the size of the network and its inherent capabilities but mostly its limited context size. All these Dutch-specific GPT models have a context window of ``only'' 512 tokens compared to 4096 of Llama 2-based models and 8192 of Mistral models. When tasks require a lot of context (e.g. 25 examples), then that may severely impact the performance of low-context models, who will cut off parts of the input that exceeds their maximum length.

Secondly, Llama~2 based models (including Orca~2) generally seem to perform worse than Mistral based models. In fact, Llama 2 13B based models perform worse than almost all Mistral-based 7B models with the exception of GEITje 7B, which performs on average as well as Orca~2 13B. Looking at the averages, these differences are relatively small, however.

Third, we see that finetuning Llama 2 on Dutch data has increased its performance but the differences are so small on average that no statistical conclusions can be drawn. Important anecdote here, however, is that while benchmarks do not show much improvement over the base model that we started from (\texttt{llama-2-13b-hf}), we do find that the model performs much better in an actual conversational manner compared to \texttt{llama-2-13b-chat-hf}, which often switches back to English. These benchmarks therefore only tell one side of the story, as we will come back to in the conclusion. Interestingly, the chat-tuned version outperforms \texttt{llama-2-13b-chat-hf} as well as \texttt{llama2-13b-ft-mc4\_nl\_cleaned\_tiny} on zero-shot TruthfulQA, but the latter model performs better in ARC and Hellaswag, which seems to suggest that the finetuning process on instructions has decreased the model's capabilities in reasoning tasks.

Finally, GEITje Chat performs really well. In fact, all models based on Mistral score highly. Interestingly, however, base GEITje performs a little bit worse, even worse than its base Mistral. Differences are so small that they should be take with a grain of salt but the performance of base GEITje on ARC is noteworthy. It seems that for reasoning, it has lost some of the qualities that base Mistral provided. GEITje Chat, on the other hand, is very competitive with Zephyr, the highest ranking model. GEITje and GEITje chat are both far ahead in the Hellaswag dataset. This is perhaps no surprise: in Hellaswag, models have to choose a sequence that most naturally follows on a give prompt. Because GEITje has been trained so thoroughly on Dutch, it likely has a better understanding and production quality of Dutch in terms of Dutch grammar and natural language completion. Due to the performance on MMLU and TruthfulQA, Zephyr is currently the leader on the board. This seems to indicate that Zephyr has a wide coverage on different tasks and topics (MMLU) and is particularly strong in truthfulness (TruthfulQA).

\begin{landscape}
\begin{table}
\centering
\begin{tabular}{lll|rrrrrr}
Model & Type & NL & Size & Avg. & ARC (25-shot) & HellaSwag (10-shot) & MMLU (5-shot) & TruthfulQA (0-shot)\\\midrule
zephyr-7b-beta & RL-tuned & none & 7.2B & \bfseries 0.49 & 0.43 & 0.58 & \bfseries 0.43 & \bfseries 0.53 \\
geitje-7b-chat & instruction-tuned & fine-tuned & 7.2B & 0.47 & 0.42 & \bfseries 0.67 & 0.33 & 0.46 \\
neural-chat-7b-v3-1 & RL-tuned & none & 7.2B & 0.47 & \bfseries 0.43 & 0.58 & 0.34 & 0.51 \\
mistral-7b-v0.1 & pretrained & none & 7.2B & 0.46 & 0.43 & 0.58 & 0.37 & 0.45 \\
orca-2-13b & instruction-tuned & none & 13.0B & 0.45 & 0.42 & 0.54 & 0.37 & 0.50 \\
geitje-7b & fine-tuned & fine-tuned & 7.2B & 0.45 & 0.38 & 0.65 & 0.32 & 0.43 \\
llama-2-13b-chat-hf & RL-tuned & none & 13.0B & 0.44 & 0.41 & 0.55 & 0.37 & 0.43 \\
llama2-13b-ft-mc4\_nl\_cleaned\_tiny & fine-tuned & fine-tuned & 13.0B & 0.44 & 0.40 & 0.58 & 0.35 & 0.42 \\
llama-2-13b-chat-dutch & instruction-tuned & fine-tuned & 13.0B & 0.43 & 0.38 & 0.56 & 0.35 & 0.44 \\
llama-2-13b-hf & pretrained & none & 13.0B & 0.43 & 0.38 & 0.57 & 0.36 & 0.41 \\
orca-2-7b & instruction-tuned & none & 6.7B & 0.41 & 0.37 & 0.49 & 0.33 & 0.45 \\
llama-2-7b-chat-hf & RL-tuned & none & 6.7B & 0.41 & 0.36 & 0.49 & 0.33 & 0.44 \\
llama-2-7b-hf & pretrained & none & 6.7B & 0.40 & 0.36 & 0.51 & 0.32 & 0.41 \\
gpt2-medium-dutch & pretrained & pretrained & 354.8M & 0.30 & 0.24 & 0.25 & 0.25 & 0.45 \\
gpt-neo-1.3b-dutch & pretrained & pretrained & 1.3B & 0.30 & 0.26 & 0.25 & 0.25 & 0.43 \\
gpt-neo-125m-dutch & pretrained & pretrained & 125.2M & 0.29 & 0.24 & 0.24 & 0.25 & 0.43 \\
gpt2-large-dutch & pretrained & pretrained & 774.0M & 0.29 & 0.24 & 0.25 & 0.24 & 0.42 \\
\end{tabular}

\caption{Benchmark results of different models on the translated-to-Dutch ARC, HellaSwag, MMLU and TruthfulQA benchmarks. \texttt{Type} indicates how the model was trained: pretrained, finetuned or preference-tuned with a form of reinforcement learning. \texttt{NL} highlights whether the model was specifically pretrained on Dutch, further fine-tuned on Dutch, or whether no specific attention was placed on including Dutch (none). All models tested in 8 bit.}
\label{tab:results}
\end{table}
\end{landscape}

\section{Discussion}
\label{sec:discussion}

In this paper we have presented language resources for Dutch language modelling, which include two Llama~2-based large language models, four synthetic datasets for instruction following and a repository to easily translate or answer other datasets, and a leaderboard for Dutch benchmarks for generative models, which we populated with results from state-of-the-art language models. 

In those benchmarks, which are automatic translations created by \cite{dac2023okapi} of existing benchmarks for English, we find that, perhaps surprisingly, scores of large language models that are not specifically trained on Dutch perform very well. In fact, the best performing model is Mistral-based Zephyr, which has not been trained specifically on Dutch. A strong contender, based on the same base model as Zephyr, is GEITje Chat, which takes the second position. The impact of the base model is of great influence on the position on the leaderboard and Mistral-based models, despite being 7B parameters, generally seem to outperform larger (13B) Llama~2 based models. In terms of size, this is an exciting prospect, as it shows that size is not everything. It begs the question what is important, then, and the answer must be mostly attributed to the training data. Unfortunately, in the current climate of large language models, qualitative datasets are worth gold, which means that it is often unknown which data these powerful models, such as Mistral, are actually trained on. In our opinion this lack of transparency, likely motivated by concerns of copyright and ethics, or maintaining a competitive edge, holds back the progress that can be made in the field.

While a large part of our paper discusses benchmark results, there are some notes to be made in this respect. We are aware that benchmarking models on translated data is not ideal. However, for Dutch there are no other options for generative models at the moment. Because the benchmarks were automatically translated, some translationese effects may occur: the translations may not be fluent Dutch or still contain artefacts of the source text (like word order, literal translation, certain vocabulary items). Because of that, an unfair advantage may be given to the non-Dutch models: Dutch is closely related to English, so if the benchmarks are in automatically translated Dutch that still has English properties, those English models may understand them better than actual, human-written, monolingual Dutch. If the benchmarks were to have been manually translated or, even better, created from scratch in Dutch, those non-Dutch models may have a harder time. We cannot know for sure until we have high-quality, manually crafted benchmarks for Dutch. Another shortcoming is that we do not calculate significance scores or confidence intervals. When results are close together in the leaderboard we therefore urge caution when interpreting the model ranks.

A second point of discussion is the relevance of these benchmarks for Dutch compared to the expected user experience. Many English benchmarks are focused on specific aspects that relate to reasoning, task understanding, and a wide coverage of topics and facts. The models are then often prompted with a multiple choice question, often guided with a few examples. Models mostly do not have to generate their own full-text answer as that would be tough to evaluate. An underlying implication, therefore, is the imbalance in evaluation between language understanding and language generation in these benchmarks, where the latter is barely evaluated. However, \textit{natural} language \textit{production} is paramount for the user experience and wide adoption of the technology. In other words, users will be turned off from a service when they notice that it does not have a good understanding of the user's request \textit{or} when the model does not generate naturally sounding answers. We believe that this discrepancy in focus in the benchmarks is caused by multiple factors. First, without manual annotation and evaluation it is difficult to evaluate the real ``naturalness'' of language generation systems. Recent methodology ends up using stronger models evaluating the output quality of smaller models, which in itself is a circular problem but perhaps a necessary evil in some fields. Second, for English there seems to be an assumption that models have become so powerful that the language that they generate \textit{is} natural so it does not warrant additional evaluation. However, when tuning a language model to a new language, this is not a given. When a model is trained on multiple languages, or first trained on one language and then finetuned for another, many linguistic interactions can take place that lead to ungrammatical language.

We remarked earlier that models such as Llama~2~13~Chat and Zephyr are powerful, and they work well in English conversation. However, when used in practice to hold a conversation in Dutch, there are a lot of linguistic artefacts that quickly break any immersion (a quick fail on the Turing test). This is especially true for the Llama~2 model but can also be clearly observed when using Zephyr for a few interactions. The models will hallucinate non-existing words, will stop producing Dutch after a few hundred tokens, will employ incorrect (morpho)syntax such as word order and plural markers, and will often read more like a poor translation from English to Dutch. This is to be expected, as the models never promise to deliver high-quality performance on Dutch. However, it does mean that the results in the benchmark should be considered specifically in what they are meant for: investigating reasoning, truthful, and task coverage, \textbf{not} language coverage nor fluency in language generation. Our finetuned Llama~2 Chat Dutch model already improves on those mostly-English baselines when it comes to interaction and user experience and especially GEITje Chat leads to a much more natural and convincing interaction. We understand that those can be considered bold claims, not justified by user experiments or thorough linguistic analysis. In future work we aim to work towards a better analysis of these experimental findings so that the naturalness and fluency of a conversation can be evaluated in a language-specific manner.

Initially, after the release of powerful generative models in the past few years, progress in generative language modelling for Dutch took off at a slow pace due to a lack of data, a lack of compute, and a lack of expertise about this new and quickly developing technology -- problems that many research groups are faced with across the globe. Since then, we have observed a rise in interest, expertise, and technological capabilities. Steps have been taken and projects are underway, the largest one to date being the GPT-NL project funded by the Dutch government.\footnote{\url{https://www.tno.nl/nl/newsroom/2023/11/nederland-start-bouw-gpt-nl-eigen-ai/}} Also on smaller scales we see more and more projects taking root with the common goal of improving the current state of Dutch generative language models. We applaud such initiatives and urge for transparency and openness, and a focus on high-quality, publicly available data and Dutch-specific, generative evaluation methodologies. Those are the required cornerstones on which we can build powerful models for our language.

\section*{Acknowledgements}

We thank the VSCentrum (\url{https://www.vscentrum.be/}, Flemish Super Computer Center) for access to its computational power. We also thank the members of Hugging Face, who have always been open to discussions and questions, and plenty of other members of the open research community who have been open to discuss ideas and research findings.

\bibliography{anthology,references}

\appendix

\section{Dataset Translation Details}
\label{app:translation}
All datasets introduced here were translated with \texttt{gpt-3.5-turbo}, with max. tokens set to $1024$ and temperature to $0$. For all tasks, the following system message was used.

\paragraph{System message}
\texttt{You are a helpful assistant that translates English to Dutch to the requirements that are given to you.}

\subsection{Dolly 15K, Alpaca Cleaned}

\paragraph{Prompt}

{\tt You are asked to translate a task's instruction, optional input to the task, and the output of the task, from English into Dutch.\\
\\
Here are the requirements that you should adhere to:\\
1. maintain the format: the task consists of a task instruction (marked `instruction: `), optional input to the task (marked `input: `) and output for the task marked with `output: `;\\
2. do not translate the identifiers `instruction: `, `input: `, and `output: ` but instead copy them to your output;\\
3. make sure that text is fluent to read and does not contain grammatical errors. Use standard Dutch without regional bias;\\
4. translate the instruction and input text using informal, but standard, language;\\
5. make sure to avoid biases (such as gender bias, grammatical bias, social bias);\\
6. if the instruction is to correct grammar mistakes or spelling mistakes then you have to generate a similar mistake in the input in Dutch, and then also generate a corrected output version in the output in Dutch;\\
7. if the instruction is to translate text from one language to another, then you do not translate the text that needs to be translated in the instruction or the input, nor the translation in the output (just copy them as-is);\\
8. do not translate code fragments but copy them to your output. If there are English examples, variable names or definitions in code fragments, keep them in English.\\
\\
Now translate the following task with the requirements set out above. Do not provide an explanation and do not add anything else.}

\subsection{Stack Overflow, Quora}

\paragraph{Prompt}

{\tt You are asked to translate a conversation between an AI assistant and a human from English into Dutch.\\
\\
Here are the requirements that you should adhere to:\\
1. maintain the format: the conversation consists of the AI (marked as `[|AI|]`) and the human (`[|Human|]`) talking in turns and responding to each other;\\
2. do not translate the speaker identifiers `[|AI|]` and `[|Human|]` but always copy them into the translation in appropriate places;\\
3. ensure accurate translation and keep the correctness of the conversation;\\
4. make sure that text is fluent to read and does not contain grammatical errors. Use standard Dutch without regional bias;\\
5. translate the human's text using informal, but standard, language;\\
6. make sure to avoid biases (such as gender bias, grammatical bias, social bias);\\
7. if the human asks to correct grammar mistakes or spelling mistakes then you have to generate a similar mistake in Dutch, and then also generate a corrected output version for the AI in Dutch;\\
8. if the human asks to translate text from one to another language, then you only translate the human's question to Dutch but you keep the translation that the AI provides in the language that the human requested;\\
9. do not translate code fragments but copy them as they are. If there are English examples, variable names or definitions in code fragments, keep them in English.\\
\\
Now translate the following conversation with the requirements set out above. Do not provide an explanation and do not add anything else.}

\end{document}